\documentclass{article}

 \usepackage[preprint]{neurips_2025}

\usepackage[utf8]{inputenc} % allow utf-8 input
\usepackage[T1]{fontenc}    % use 8-bit T1 fonts
\usepackage{hyperref}       % hyperlinks
\usepackage{url}            % simple URL typesetting
\usepackage{booktabs}       % professional-quality tables
\usepackage{amsfonts}       % blackboard math symbols
\usepackage{nicefrac}       % compact symbols for 1/2, etc.
\usepackage{microtype}      % microtypography
\usepackage{xcolor}         % colors

\usepackage[accsupp]{axessibility}

\usepackage{graphicx}
\usepackage{amsmath}
\usepackage{amssymb}
\usepackage{amsmath, amssymb}
\usepackage{bm}

\usepackage{booktabs}
\usepackage{multirow}
\usepackage{float}
\usepackage[labelformat=simple, labelfont=small, labelsep=period]{subcaption}

\usepackage[capitalize]{cleveref}
\crefname{section}{Sec.}{Secs.}
\Crefname{section}{Section}{Sections}
\Crefname{table}{Table}{Tables}
\crefname{table}{Tab.}{Tabs.}

\usepackage{xcolor}

\newenvironment{customlegend}[1][]{%
    \begingroup
    \csname pgfplots@init@cleared@structures\endcsname
    \pgfplotsset{#1}%
}{%
    \csname pgfplots@createlegend\endcsname
    \endgroup
}%

\def\addlegendimage{\csname pgfplots@addlegendimage\endcsname}
\definecolor{acolor}{rgb}{1.0, 0.5490196078431373, 0.0}
\definecolor{bcolor}{rgb}{0.5019607843137255, 0.0, 0.5019607843137255}
\definecolor{ccolor}{rgb}{0.0, 0.5019607843137255, 0.0}
\definecolor{darkorange}{rgb}{1.0, 0.55, 0.0} % same as matplotlib darkorange

\definecolor{dcolor}{rgb}{0, 0.302, 0.596}
\definecolor{ecolor}{rgb}{0.647, 0, 0.267}
\definecolor{fcolor}{rgb}{0, 0, 0}

\usepackage{caption}
\usepackage{subcaption}
\usepackage{tikz}
\usepackage{pgfplots}
\usepgfplotslibrary{fillbetween}
\pgfplotsset{compat=1.16}
\usepackage{multicol}

\title{CoPE: A Lightweight Complex Positional Encoding}

\author{Avinash Amballa \\
University of Massachusetts Amherst, USA\\
{aamballa@umass.edu}
}

\begin{document}

\maketitle

\begin{abstract}
 Recent studies have demonstrated the effectiveness of position encoding in transformer architectures. By incorporating positional information, this approach provides essential guidance for modeling dependencies between elements across different sequence positions. We introduce CoPE (a lightweight Complex Positional Encoding), a novel architecture that leverages complex-valued encoding to encode both content and positional information. Our approach replaces traditional positional encodings with complex embeddings where the real part captures semantic content and the imaginary part encodes positional information. We introduce phase-aware attention in the first layer of the transformer model to capture position-dependent patterns, followed by standard attention layers for higher-levels. We show that CoPE doesn't exhibit long term decay and is compatible with  linear attention. Experimental evaluation on the GLUE benchmark suggest that our approach achieves superior performance with less computational complexity, compared to RoPE, Sinusoidal and Learned positional encodings. 
\end{abstract}

\section{Introduction}

The sequential order of words plays a crucial role in natural language understanding. Traditional approaches, such as recurrent neural networks (RNNs), model word order by recursively updating hidden states over time. The Transformer architecture \cite{vaswani2023attentionneed}  has fundamentally transformed the landscape of natural language processing and sequence modeling since its introduction. While the self-attention mechanism enables the model to capture long-range dependencies without the sequential constraints of recurrent architectures, it inherently lacks positional awareness. This limitation necessitates explicit positional encoding mechanisms to inform the model about token positions within sequences. 

To address this, researchers have proposed multiple strategies for integrating positional information into the learning process. Traditional approaches employ additive positional encodings, where sinusoidal  \cite{vaswani2023attentionneed} or learned positional vectors \cite{gehring2017convolutionalsequencesequencelearning, devlin2019bertpretrainingdeepbidirectional,lan2020albertlitebertselfsupervised, Radford2018ImprovingLU} are element-wise added to token embeddings before being fed into attention layers. On the other hand \cite{su2023roformerenhancedtransformerrotary, dai2019transformerxlattentivelanguagemodels, Raffel2019ExploringTL, shaw2018selfattentionrelativepositionrepresentations} proposed relative Position Encoding which encodes the relative position information into the attention mechanism. 

The conventional additive approach to positional encoding, while effective, presents several theoretical and practical limitations. When positional information is directly added to semantic embeddings, it leads to information interference, where the model struggles to disentangle positional and semantic information. This interference becomes particularly problematic in tasks requiring precise positional reasoning. In addition, its effectiveness diminishes when applied to longer sequences \cite{shaw2018selfattentionrelativepositionrepresentations, liutkus2021relativepositionalencodingtransformers}. On the other hand, most of the relative postional encodings are unsuitable for linear self-attention architecture as shown in RoPE \cite{su2023roformerenhancedtransformerrotary} Moreover, these encodings inherently enforce long-term decay, a suboptimal inductive bias, given that modern LLMs frequently require access to information from arbitrary context positions \cite{chen2024hopenovelpositionalencoding}.

Building upon these insights, we propose CoPE, a novel light weight complex positional encoding that leverages complex-valued encoding to encode positional information through phase components while preserving semantic content in the magnitude. Our approach fundamentally reimagines positional encoding by utilizing the natural separation between real and imaginary components of complex numbers, thereby avoiding the interference inherent in additive methods. To this end, we introduce phase-aware attention in the first layer to capture the complex positional encoding and retain the standard attention in the rest layers, making it a lightweight adapter than can be integrated to any existing models. We also show that our method is compatible with linear attention and doesn't exhibit long term decay. We evaluate our method on several GLUE benchmarks and results suggest that CoPE is superior to RoPE, Sinusoidal and Learned encodings. \\

To summarize, our key contributions are:
\begin{enumerate}
     \item  A novel lightweight complex encoding that separates semantic content (real part) and positional information (imaginary part)
    \item A phase-aware attention mechanism in the first layer that leverages both magnitude and phase information. We introduce several types of phase-aware attention. 
     \item We show that CoPE doesn't exhibit long term decay and is compatible with Linear attention.
     \item  We evaluate our method on various GLUE benchmarks, and results suggest that CoPE achieves superior performance compared to its alternatives. 
\end{enumerate}

\section{Related work} 

\subsection{Positional Encoding}

\textbf{Absolute} positional encodings focus on individual position information and is
typically applied in the first layer of the model. These are embeddings that are directly added to the input token embeddings. Sinusoidal positional encodings \cite{vaswani2023attentionneed} are non-learned vectors that are added directly to input embeddings to at the bottoms of the encoder and decoder stacks. On the other hand, Learned encodings \cite{wang2020position} use a learned additive vector. 

\textbf{Relative} positional encoding focus on relative position rather than absolute. While absolute positional encoding (APE) offers a straightforward and intuitive approach, its effectiveness diminishes when applied to longer sequences. This limitation has led researchers to increasingly focus on refining relative positional encoding. \cite{shaw2018selfattentionrelativepositionrepresentations, liutkus2021relativepositionalencodingtransformers} use relative position encodings that attempts to exploit pairwise, relative positional information. Relative positional information is supplied to the model on two levels: values and keys.  RoPE \cite{su2023roformerenhancedtransformerrotary} integrates position awareness throughout all transformer layers. Notably, this method uniquely preserves the position-agnostic nature of the value vectors in the self-attention mechanism.

\textbf{Additive relative positional} encoding introduce bias matrix to the attention matrix. 
Its popular variants include  T5 Bias \cite{Raffel2019ExploringTL}, ALiBi \cite{press2022trainshorttestlong}. ALiBi penalizes the attention value so that a query can assign to the key depending on how far away the key and query are. So when a key and query are close by, the penalty is very low, and when they are far away, the penalty is very high. It outperforms those methods and Rotary embeddings when evaluating sequences that are longer than the ones the model was trained on (extrapolation).

Other positional embeddings include conditional positonal encoding \cite{chu2023conditionalpositionalencodingsvision}
 which  generate and conditions on the local neighborhood of the input tokens. \cite{hua2025fourierpositionembeddingenhancing} introduces Fourier positional embedding which enhances attention's frequency-domain properties to improve both its periodic extension and length generalization. HoPE \cite{chen2024hopenovelpositionalencoding} replaces the specific components in RoPE with
position-independent ones, retaining only high frequency signals, leading to greater
robustness to the out-of-distribution.

\subsection{Complex-Valued Neural Networks}
Complex-valued neural networks \cite{9849162, bassey2021surveycomplexvaluedneuralnetworks} have emerged as a promising paradigm for processing information with inherent multi-dimensional structure. 
Unlike real-valued networks that operate on scalar features, complex networks naturally encode information through both magnitude and phase components, providing orthogonal dimensions for representation. This dual-component structure has proven particularly effective in signal processing applications \cite{bassey2021surveycomplexvaluedneuralnetworks} where phase information carries crucial meaning. 

\cite{Eilers_2023} introduce complex-valued neural networks by presenting building blocks to transfer the transformer architecture to the complex domain. They present multiple versions of a complex-valued Scaled Dot-Product Attention mechanism as well as a complex-valued layer normalization. \cite{leng2025unveilingpowercomplexvaluedtransformers} propose fundamental paradigm of complex-valued transformers for wireless communications.  

\section{Method}

\subsection{Complex Encoding Layer}

Our work is closely related to \cite{wang2020encodingwordordercomplex} which extends word vectors as continuous functions over changing variables like word position. They also introduce a general complex-valued word embedding approach where each word-position combination is represented as a waveform with trainable amplitude ($r_j$), frequency ($\omega_j$), and phase ($\theta_j$) parameters. Unlike traditional additive position embeddings, this method uses element-wise multiplication between word embeddings and positional components, allowing adaptive control over position sensitivity per dimension. 

In this work, we isolate position into the complex domain, keeping token embeddings real. Our approach begins with complex-valued encoding that encodes content and position separately. In particular, 
% $
% E_{complex}(x, pos) = E_{vocab}(x) + i · E_{pos}(pos)
% $
\[
E_{\text{complex}}(x, \text{pos}) = E_{\text{vocab}}(x) + i \cdot E_{\text{pos}}(\text{pos})
\]

where $E_{\text{vocab}}(x)$ represents the token embedding (real part) \footnote{we add the sentence embedding to the token embedding if applicable}, $E_{\text{pos}}(\text{pos}$) represents the positional embedding (imaginary part). Here $i$ represents the imaginary unit

This representation naturally separates semantic content from positional information while maintaining their relationship through the complex structure. In this paper, we use sinusoidal encoding \cite{vaswani2023attentionneed} in imaginary part to encode position information to extrapolate beyond the trained sequence length. 
\[
E_{\text{complex}}(x, \text{pos}) = E_{\text{vocab}}(x) + i \cdot \gamma \cdot \sin{(\omega \cdot pos)}
\]

% $
% E_{complex}(x, pos) = E_{vocab}(x) + i · \gamma \cdot \sin{(\omega \cdot pos)}
% $

Complex representations offer theoretical advantages for positional modeling. 

\begin{enumerate}
    \item  Orthogonal Information Encoding: Real and imaginary components are orthogonal, preventing direct interference between content and position. Unlike additive positional encodings that cause information loss through vector addition, complex embeddings preserve both components. 
    \item Rotation: Complex multiplications \cite{Eilers_2023} enables position-dependent transformations through rotation. 
\end{enumerate}

\subsection{Phase-Aware Attention}

To model the complex input from positional information, we introduce phase-aware attention mechanism in the first layer. This attention mechanism captures both semantic and positional relationships. 

To handle the complex valued embedding, we introduce complex valued projection in the first layer. Let the complex-valued projections be defined as:
\[
Q_{\text{proj}} = Q_{\text{real}} + i \cdot Q_{\text{imag}}, \quad 
K_{\text{proj}} = K_{\text{real}} + i \cdot K_{\text{imag}}
\]

For complex input, $z = z_{\text{real}} + i \cdot z_{\text{imag}}$, the complex-valued query and key vectors are:

\[
\begin{aligned}
Q_{\text{complex}} &= Q_{\text{proj}} \cdot z \\
&= (Q_{\text{real}} \cdot z_{\text{real}} - Q_{\text{imag}} \cdot z_{\text{imag}}) \\
&    + i · (Q_{\text{real}} \cdot z_{\text{imag}} + Q_{\text{imag}} \cdot z_{\text{real}})
\end{aligned}
\begin{aligned}
K_{\text{complex}} &= K_{\text{proj}} \cdot z \\
&= (K_{\text{real}} \cdot z_{\text{real}} - K_{\text{imag}} \cdot z_{\text{imag}}) \\
&   + i (K_{\text{real}} \cdot z_{\text{imag}} + K_{\text{imag}} \cdot z_{\text{real}})
\end{aligned}
\]

We keep the value vector $V$ in real space (projection on $z_{real}$), to propogate the real valued output to next layers. 

We define the attention scores in a similar fashion to \cite{Eilers_2023} to incorporate both magnitude and phase information in the attention computation:

\[
A_{\text{complex}} = Q_{\text{complex}} \cdot K_{\text{complex}}^*
\]
Here $^*$ denotes the complex conjugate. Let $A_{\text{magnitude}}$, $A_{\text{phase}}$, and $\Re(A_{\text{complex}})$ represent the magnitude, phase, and real part of the complex attention scores $A_{\text{complex}}$, respectively.

We propose several variants to map the complex-valued attention scores to real-valued scores:
\label{attebtion_real}
\begin{enumerate}
    \item \textbf{Magnitude:}
    \[
    A_{\text{real}} = \frac{A_{\text{magnitude}}}{\sqrt{d_k}}
    \]
    
    \item \textbf{Phase:}
    \[
    A_{\text{real}} = \frac{\cos(A_{\text{phase}})}{\sqrt{d_k}}
    \]
    
    \item \textbf{Real:}
    \[
    A_{\text{real}} = \frac{\Re(A_{\text{scores}})}{\sqrt{d_k}}
    \]
    
    \item \textbf{Hybrid:}
    \[
    A_{\text{real}} = \frac{\left( A_{\text{magnitude}} + \alpha \cdot \cos(A_{\text{phase}}) \right)} {\sqrt{d_k}}
    \]
    
    \item \textbf{Hybrid-norm:}
    \[
    A_{\text{real}} = \frac{\frac{A_{\text{magnitude}}}{\max(A_{\text{magnitude}})} + \alpha \cdot \cos(A_{\text{phase}})}{\sqrt{d_k}}
    \]
\end{enumerate}

Here $\alpha$ is a phase coefficient controlling phase influence. We choose cosine function in phase to model similarity i.e. lesser the phase difference, the more the similar as discussed in \cite{Eilers_2023}

Since attention score  $A_{\text{real}}$ $\&$ value vectors $V$ are still in real valued space, we use $Softmax(A_{\text{real}})*V$ 

\subsection{Properties of CoPE}

\textbf{1. CoPE doesn't exhibit Long term decay}: 
Recent work on HoPE \cite{chen2024hopenovelpositionalencoding} challenges the conventional assumption that positional encodings must enforce long-term decay, arguing that modern LLMs often need to retrieve information from arbitrary context positions. In this section, we prove that CoPE doesn't exhibit long term decay. 

Reformulating our definition of the complex positional embedding for token $x$ at position $p$ as
\begin{equation}
    \bm{z}(x, p) = \bm{e}_x + i \, \gamma \, \sin(\omega p),
\end{equation}
where $\bm{e}_x \in \mathbb{R}^d$ is the token embedding, $\gamma \in \mathbb{R}$ is a scaling factor, and $\omega$ is the base angular frequency.

We consider the complex inner product:
\begin{equation}
    A_{\text{complex}}(x,y,p,q) = Q_{\text{complex}}(x, p) \cdot K_{\text{complex}}(y, q)^{*}.
\end{equation}

Substituting the definitions:
$    A_{\text{complex}}(x,y,p,q) $ \\
\begin{align}
    = \left( Q_{\text{proj}} \left[ \bm{e}_x + i \, \gamma \, \sin(\omega p) \right] \right) 
       \cdot 
       \left(  K_{\text{proj}} \left[ \bm{e}_y + i \, \gamma \, \sin(\omega q) \right] \right)^{*} \\
    = \left( Q_{\text{proj}} \bm{e}_x + i \gamma Q_{\text{proj}} \sin(\omega p) \right)
       \cdot
       \left(  K_{\text{proj}} \bm{e}_y - i \gamma  K_{\text{proj}} \sin(\omega q) \right) \\
    = \underbrace{\left( Q_{\text{proj}} \bm{e}_x \right) \cdot \left(  K_{\text{proj}} \bm{e}_y \right)}_{\text{content term}}
    + i \gamma \left[ \left( Q_{\text{proj}} \bm{e}_x \right) \cdot \left( -  K_{\text{proj}} \sin(\omega q) \right) \right] \\
    \quad + i \gamma \left[ \left( Q_{\text{proj}} \sin(\omega p) \right) \cdot \left(  K_{\text{proj}} \bm{e}_y \right) \right] \\
     + \underbrace{\gamma^2 \left[ Q_{\text{proj}} \sin(\omega p) \cdot  K_{\text{proj}} \sin(\omega q) \right]}_{\text{position term}}.
\end{align}

The last term encodes the positional interaction:
\begin{align}
    Q_{\text{proj}} \sin(\omega p) \cdot  K_{\text{proj}} \sin(\omega q) 
    \propto \sin(\omega p) \sin(\omega q) \\
    = \frac{1}{2} \left[ \cos(\omega(p - q)) - \cos(\omega(p + q)) \right].
\end{align}

Thus, the positional contribution to $A_{\text{complex}}(x,y,p,q)$ is
\begin{equation}
    A_{\text{complex}}(x,y,p,q)  \propto \cos\left( \omega (p - q) \right) - \cos\left( \omega (p + q) \right) \label{eq_phase}
\end{equation}

The relative position term $\cos(\omega (p - q))$ is \emph{purely oscillatory} with respect to $p - q$ and has no multiplicative decay factor such as $e^{-\alpha |p-q|}$. Therefore, this complex encoding does \textbf{not} impose long-term decay on the attention score magnitude.
\\

\textbf{2. CoPE encodes both relative and absolute positions}: We note that CoPE embeddings are absolute. However, phase-aware attention encodes both relative and absolute positions. The relative position information emerges naturally from the phase difference encoded via complex multiplication i.e., $A_{\text{complex}} = Q_\text{complex}  K_\text{complex}^{*}$. Given positions $p$, 
q, $A_{\text{complex}} \propto \cos\left( \omega (p - q) \right) - \cos\left( \omega (p + q) \right).$ as shown in eq \ref{eq_phase}
\\

\textbf{3. CoPE is compatible with Linear Attention}:
We show that CoPE with phase-aware attention is compatible with linear attention mechanisms. Linear attention \cite {Katharopoulos2020TransformersAR} rewrites the attention as:
\[
\text{Attention}(Q, K, V)_m = \frac{ \sum_{n=1}^N \phi(q_m)^\top \phi(k_n) v_n }{ \sum_{n=1}^N \phi(q_m)^\top \phi(k_n) },
\]
where $\phi(x)$ is a non-negative activation function such as elu(x) + 1

To incorporate complex queries and keys, we lift the complex vectors to doubled real features by splitting real, imaginary parts and applying \(\phi\) separately:
\[
\phi(q) \;=\; \begin{bmatrix} \phi(q_r) \\[2pt] \phi(q_i) \end{bmatrix}
\in\mathbb{R}^{2d}, \qquad
\phi(k) \;=\; \begin{bmatrix} \phi(k_r) \\[2pt] \phi(k_i) \end{bmatrix}
\in\mathbb{R}^{2d}.
\]

We compute the Hermitian inner product for the lifted features:
\begin{align}
\phi(q)^\dagger \phi(k)
&= \big(\phi(q_r) - i\phi(q_i)\big)^\top \big(\phi(k_r) + i\phi(k_i)\big) \\
&= \big(\phi(q_r)^\top\phi(k_r) + \phi(q_i)^\top\phi(k_i)\big) \\
&\;+\; i\big(\phi(q_i)^\top\phi(k_r) - \phi(q_r)^\top\phi(k_i)\big).
\label{eq:kernel_expand}
\end{align}
Thus the complex kernel decomposes into four real inner products:
\[
A_{rr}=\phi(q_r)^\top\phi(k_r),\quad
A_{ii}=\phi(q_i)^\top\phi(k_i)\quad
\]

\[
A_{ir}=\phi(q_i)^\top\phi(k_r),\quad
A_{ri}=\phi(q_r)^\top\phi(k_i).
\]

Plugging \(\phi\) into the linear-attention numerator gives
\begin{align}
\mathrm{Num}_m
&= \sum_{n=1}^N \phi(q_m)^\dagger \phi(k_n)\, v_n \\
&= \sum_{n=1}^N \Big( (A_{rr}^{(m,n)}+A_{ii}^{(m,n)}) + i (A_{ir}^{(m,n)}-A_{ri}^{(m,n)}) \Big) v_n.
\end{align}
Each real inner product \(A_{uv}^{(m,n)}=\phi(Q_{u,m})^\top\phi(K_{v,n})\) is separable in \(m\) and \(n\). Therefore we can precompute key–value aggregates:
\begin{align}
\mathbf{G}_{r} &:= \sum_{n=1}^N \phi(k_{r,n})\, v_n^\top, \qquad
\mathbf{G}_{i} := \sum_{n=1}^N \phi(k_{i,n})\, v_n^\top \\
\end{align}

Thus the numerator is computed by a small number of matrix–vector products:
\begin{align}
\mathrm{Num}_m
&= \big( \phi(q_{r,m})^\top \mathbf{G}_{r} + \phi(q_{i,m})^\top \mathbf{G}_{i} \big) \\
& \;+\; i\big( \phi(q_{i,m})^\top \mathbf{G}_{r} - \phi(q_{r,m})^\top \mathbf{G}_{i} \big).
\label{eq:num_computed}
\end{align}

For numerical stability and to avoid division by complex scalars, a common choice is to keep the denominator real. For instance one may set
\[
\mathrm{Den}_m \;=\; \sum_{n=1}^N \big( \phi(q_{r,m})^\top \phi(k_{r,n}) + \phi(q_{i,m})^\top \phi(k_{i,n}) \big),
\]

% --- Figure 1: SST2 ---
\begin{figure}[t]
    \centering
    \includegraphics[width=0.8\linewidth]{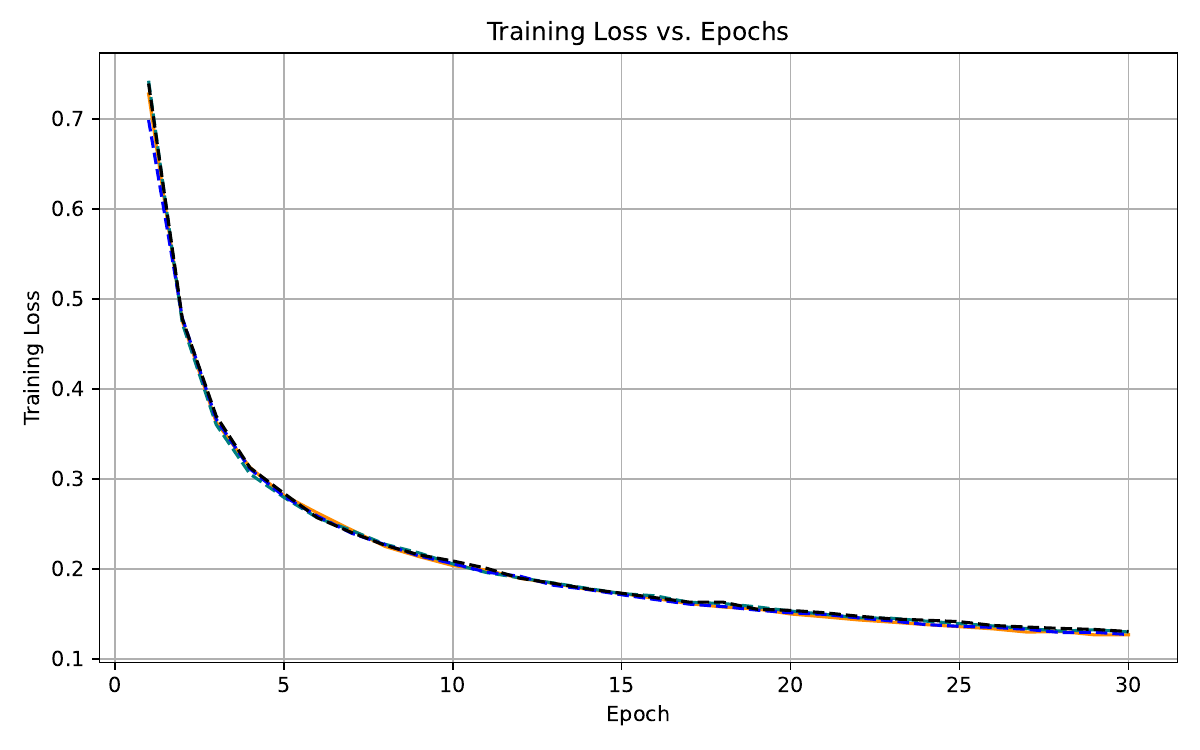}
    \caption{Training loss vs. epochs on SST2 with CoPE vs RoPE.}
    \label{fig:sst}
    
    % Legend
    \vspace{0.5em}
    \begin{tikzpicture}
    \begin{customlegend}[
        legend columns=4,
        legend style={
            draw=black,
            fill=white,
            inner sep=2pt,
            font=\small
        },
        legend entries={
            Complex Valued phase,
            Complex Valued magnitude,
            Complex Valued hybrid norm,
            ROPE
        }
    ]
        \addlegendimage{color=teal,solid,dashed}
        \addlegendimage{color=blue,solid,dashed}
        \addlegendimage{color=black,solid,dashed}
        \addlegendimage{color=darkorange,solid}
    \end{customlegend}
    \end{tikzpicture}
\end{figure}

Given the complex numerator \(\mathrm{Num}_m \in \mathbb{C}^{V}\) and real denominator \(\mathrm{Den}_m \neq 0\), one can form different real-valued attention outputs as shown in sec \ref{attebtion_real}:
\begin{itemize}
  \item \textbf{Magnitude:} \(\mathrm{Attention}_m = |\mathrm{Num}_m|/\mathrm{Den}_m.\)
  \item \textbf{Phase:} \(\mathrm{Attention}_m =\ cos(\arg(\mathrm{Num}_m))/\mathrm{Den}_m.\)
  \item \textbf{Real:} \(\mathrm{Attention}_m = \operatorname{Re}(\mathrm{Num}_m)/\mathrm{Den}_m.\)
  \item \textbf{Hybrid:} \(\mathrm{Attention}_m = \big(|\mathrm{Num}_m| + \alpha\cos(\arg(\mathrm{Num}_m))\big)/\mathrm{Den}_m.\)
\end{itemize}
All these options use the precomputed aggregates and therefore retain \(O(N)\) complexity.

\subsection{Computation cost}

Our method applies phase-aware attention only to the first layer, followed by standard attention layers. This design captures position-dependent patterns early while allowing higher layers to focus on semantic relationships. In addition, limiting complex operations to one layer makes this encoding easy to adapt and maintains reasonable computational cost.

In this section, we show the compute cost for CoPE vs RoPE. Define number of layer in model to be $L$, number of test data samples be $N$, input sequence length $T$, number of heads $H$, $d_{k} = d_{model}/H$. RoPE \cite{su2023roformerenhancedtransformerrotary} rotates query and key vectors in every layer for every head i.e, matrix rotations complexity is $O(L*c_{rot}*N*H*T*d_k),$ assuming the rotation operations are worth $c_{rot}$. However, CoPE use phase-aware attention in single layer and standard attention in rest of the layers making the complex-valued operations complexity to be $O(c_{complex}*N*H*T*d_k)$. Ignoring the rotation factor $c_{rot}$ in RoPE and complex operation factor $c_{complex}$ in CoPE, CoPE is $L$ times faster than RoPE.

% --- Figure 2: MRPC ---
\begin{figure}[t]
    \centering
    \includegraphics[width=0.8\linewidth]{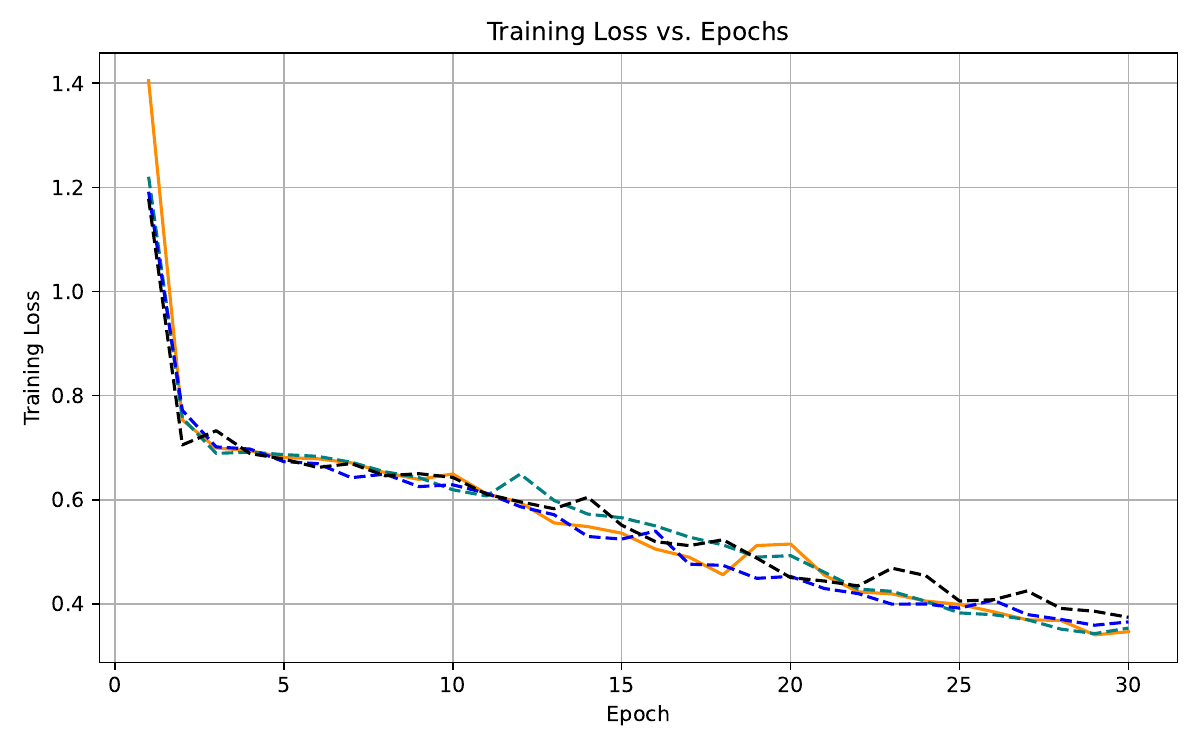}
    \caption{Training loss vs. epochs on MRPC with  CoPE vs RoPE.}
    \label{fig:mrpc}
    
    % Legend
    \vspace{0.5em}
    \begin{tikzpicture}
    \begin{customlegend}[
        legend columns=4,
        legend style={
            draw=black,
            fill=white,
            inner sep=2pt,
            font=\small
        },
        legend entries={
            Complex Valued phase,
            Complex Valued magnitude,
            Complex Valued hybrid norm,
            ROPE
        }
    ]
        \addlegendimage{color=teal,solid,dashed}
        \addlegendimage{color=blue,solid,dashed}
        \addlegendimage{color=black,solid,dashed}
        \addlegendimage{color=darkorange,solid}
    \end{customlegend}
    \end{tikzpicture}
\end{figure}

% --- Figure 3: QNLI ---
\begin{figure}[t]
    \centering
    \includegraphics[width=0.8\linewidth]{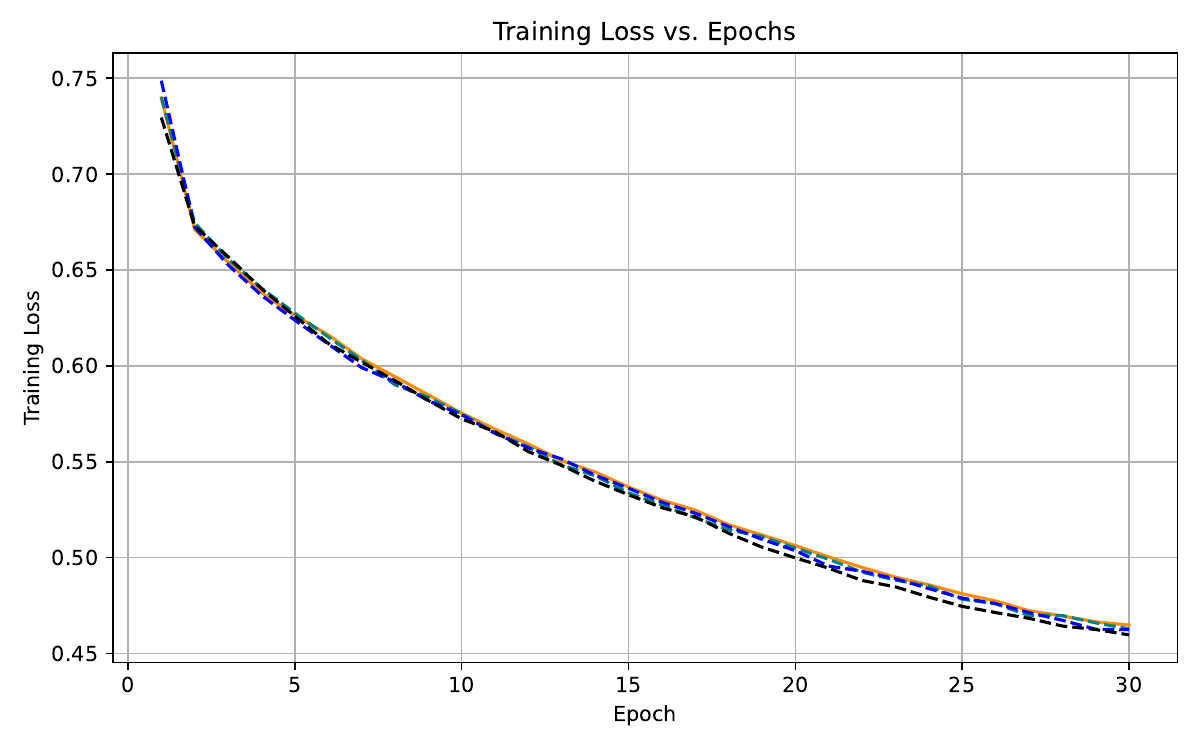}
    \caption{Training loss vs. epochs on QNLI with  CoPE vs RoPE.}
    \label{fig:qnli}
    
    % Legend
    \vspace{0.5em}
    \begin{tikzpicture}
    \begin{customlegend}[
        legend columns=4,
        legend style={
            draw=black,
            fill=white,
            inner sep=2pt,
            font=\small
        },
        legend entries={
            Complex Valued phase,
            Complex Valued magnitude,
            Complex Valued hybrid norm,
            ROPE
        }
    ]
        \addlegendimage{color=teal,solid,dashed}
        \addlegendimage{color=blue,solid,dashed}
        \addlegendimage{color=black,solid,dashed}
        \addlegendimage{color=darkorange,solid}
    \end{customlegend}
    \end{tikzpicture}
\end{figure}

\begin{table*}[t]
\centering
\begin{tabular}{lccc}
\toprule
\textbf{Encoding} & \textbf{SST2 (Accuracy)} & \textbf{MRPC (F1)} & \textbf{QNLI (Accuracy)} \\
\midrule
Learned \cite{wang2020position} & \underline{81.54} & \underline{81.55} & 60.21 \\
Sinusoidal \cite{vaswani2023attentionneed} & \textbf{82.57} & 79.74 & \textbf{63.87} \\
RoPE \cite{su2023roformerenhancedtransformerrotary} & 81.31 & 80.98 & 60.74 \\
CoPE magnitude (Ours) & 80.28 & 80.19 & \underline{61.63} \\
CoPE phase (Ours) & \textbf{82.57} & \textbf{81.71} & 59.86 \\
CoPE real (Ours) & 80.50 & 80.88 & 60.97 \\
CoPE hybrid (Ours) & 81.31 & 79.75 & 60.74 \\
CoPE hybrid-norm (Ours) & 79.13 & 81.00 & 60.74 \\
\bottomrule
\end{tabular}
\caption{Test Performance comparison of different positional encodings across multiple datasets. \textbf{Bold} and \underline{underline} indicate the best and second-best result in each column.}
\label{tab:all_results}
\end{table*}

\section{Experiments}

\subsection{Experimental Setup}

\textbf{Dataset}:
We experiment with several datasets from GLUE, i.e. MRPC \cite{Dolan2005AutomaticallyCA}, SST-2 \cite{Socher2013RecursiveDM}, QNLI for training tasks. 

\begin{enumerate}
    \item MRPC: The Microsoft Research Paraphrase Corpus consists of sentence pairs and evaluates whether two sentences are semantically equivalent. 
    \item SST2: The Stanford Sentiment Treebank (binary classification version) uses single movie review sentences to assess sentiment. 
    \item QNLI: The Question-answering Natural Language Inference dataset is a question paired with a sentence from a passage. The task is to determine if the sentence contains the answer to the question.
\end{enumerate}

\textbf{Metrics}: We use the same evaluation metrics as in RoPE \cite{su2023roformerenhancedtransformerrotary} i.e., F1-score for MRPC, and accuracy for the remaining as the evaluation metrics.

\textbf{Model Configuration}: We use transfomer model with 6 layers, 8 heads, 256-dimensional embeddings, 256-dimensional attention with max positions 512.

\textbf{Training details}: We use AdamW optimizer with learning rate 1e-4, 0.01 weight decay, and dropout $0.2$. All the model are trained from scratch for 30 epochs. We set $\alpha = 0.2$, $\gamma = 1$.

\subsection{Results} \label{results}

To visualize the training performance of CoPE, we plot the training loss vs. number of epochs for different variant of CoPE i.e., CoPE magnitude, CoPE phase, CoPE hybrid-norm and compare with RoPE. Figure \ref{fig:sst}, \ref{fig:mrpc}, \ref{fig:qnli} depicts the training loss vs number of epochs for different variants of CoPE and RoPE on SST2, MRPC, QNLI datasets respectively. On SST2, Figure \ref{fig:sst} depicts that the training loss of CoPE isclosely mirrors that of RoPE which implies that CoPE is comparable with RoPE. On MRPC and QNLI,  Figures \ref{fig:mrpc}, \ref{fig:qnli} show that CoPE achieves lower training loss compared to RoPE, in particular CoPE magnitude, CoPE phase outperforms all its competitors, indicating a more effective learning process.

These training trends are reflected in the final test performance, detailed in Table \ref{tab:all_results}. These results shows the test performance (accuracy, F1 score) of different encoding methods on SST2, MRPC, QNLI datasets. On SST2 $\&$ MRPC datasets, CoPE phase outperforms all the existing positional encodings including RoPE. On QNLI, CoPE magnitude achieves the second best performance after sinusoidal encodings, outperforming RoPE and learned positional encoding. These results demonstrate that our complex positional encoding with phase-aware attention achieves superior performance on different GLUE benchmarks with less computational complexity compared to RoPE.

% For detailed comparision of RoPE, sinusoidal and learned encodings, we encourage readers to refer to original papers. 

\section{Limitations}

\textbf{1. Extrapolation:}

Our method also allows to extrapolate beyond the sequence length due to the sinusoidal embeddings in complex domain. However, AliBi \cite{press2022trainshorttestlong} show that sinusoidal embeddings underperform when extrapolated beyond sequence length. We plan to include the extrapolation experiments with CoPE and compare with AliBi \cite{press2022trainshorttestlong}.

\textbf{2. Pretraining $\&$ Finetuning tasks}

Due to resource constraints, our current method is only evaluated on relatively smaller model that is trained from scratch. In particular, CoPE requires a separate evaluation on pretraining and fine tuning tasks on larger models.

\section{Conclusion}

We introduce CoPE, a novel positional encoding that encodes context and position information through real and imaginary components respectively. Our approach demonstrates that the application of phase-aware attention on the first layer can effectively capture positional dependencies while maintaining computational efficiency. We show that CoPE doesn't exhibit long term decay and is compatible with linear attention, make it a light weight adapter to existing models. Experimental results on the GLUE benchmarks demonstrate our approach, achieving superior performance compared to its alternatives. This work opens several research directions, including emphasis of complex space in transformers and its applications to diverse sequence modeling tasks.

\clearpage

%%%%%%%%% REFERENCES
{\small
\bibliographystyle{plainnat}
\bibliography{egbib}
}

\clearpage

\end{document}